\documentclass[10pt,twocolumn,letterpaper]{article}

\usepackage[pagenumbers]{cvpr}      
\definecolor{cvprblue}{rgb}{0.21,0.49,0.74}
\usepackage[pagebackref,breaklinks,colorlinks,allcolors=cvprblue]{hyperref}
\usepackage{multirow}
\title{Rep-GLS: Report-Guided Generalized Label Smoothing for Robust Disease Detection}

\author{Kunyu Zhang$^1$\quad
Fukang Ge$^2$\quad
Binyang Wang$^3$\quad
Yingke Chen$^4$\\
Kazuma Kobayashi$^5$\quad
Lin Gu$^6$\quad
Jinhao Bi$^7$\quad
Yingying Zhu$^{2*}$\\[1ex]
$^1$Arizona State University\quad \\
$^2$Guangzhou Institutes of Biomedicine and Health, Chinese Academy of Sciences\\
$^3$Kunming Medical University\quad
$^4$Northumbria University\\
$^5$NII, Tokyo Institute of Technology\quad
$^6$RIKEN\\
$^7$Westlake University\\[1ex]
{\tt\small zhu\_yingying@gibh.ac.cn}$^*$
}

\begin{document}
\maketitle
\begin{abstract}
Unlike nature image classification where groundtruth label is explicit and of no doubt, physicians commonly interpret medical image conditioned on certainty like using phrase "probable" or "likely". Existing medical image datasets either simply overlooked the nuance and  polarise into binary label. Here, we propose a novel framework that leverages a Large Language Model (LLM) to directly mine medical reports to utilise the uncertainty relevant  expression for supervision signal. At first, we collect uncertainty keywords from medical reports. Then, we use Qwen-3 4B to identify the textual uncertainty  and map them  into an adaptive Generalized Label Smoothing (GLS) rate. This rate allows our model to treat uncertain labels not as errors, but as informative signals, effectively incorporating expert skepticism into the training process. We establish a new clinical expert uncertainty-aware benchmark to rigorously evaluate this problem. Experiments demonstrate that our approach significantly outperforms state-of-the-art methods in medical disease detection. The curated uncertainty words database, code, and benchmark will be made publicly available upon acceptance.
\end{abstract}


\section{Introduction}

\begin{figure*}[!t]
    \centering
    \includegraphics[width=0.9\textwidth]{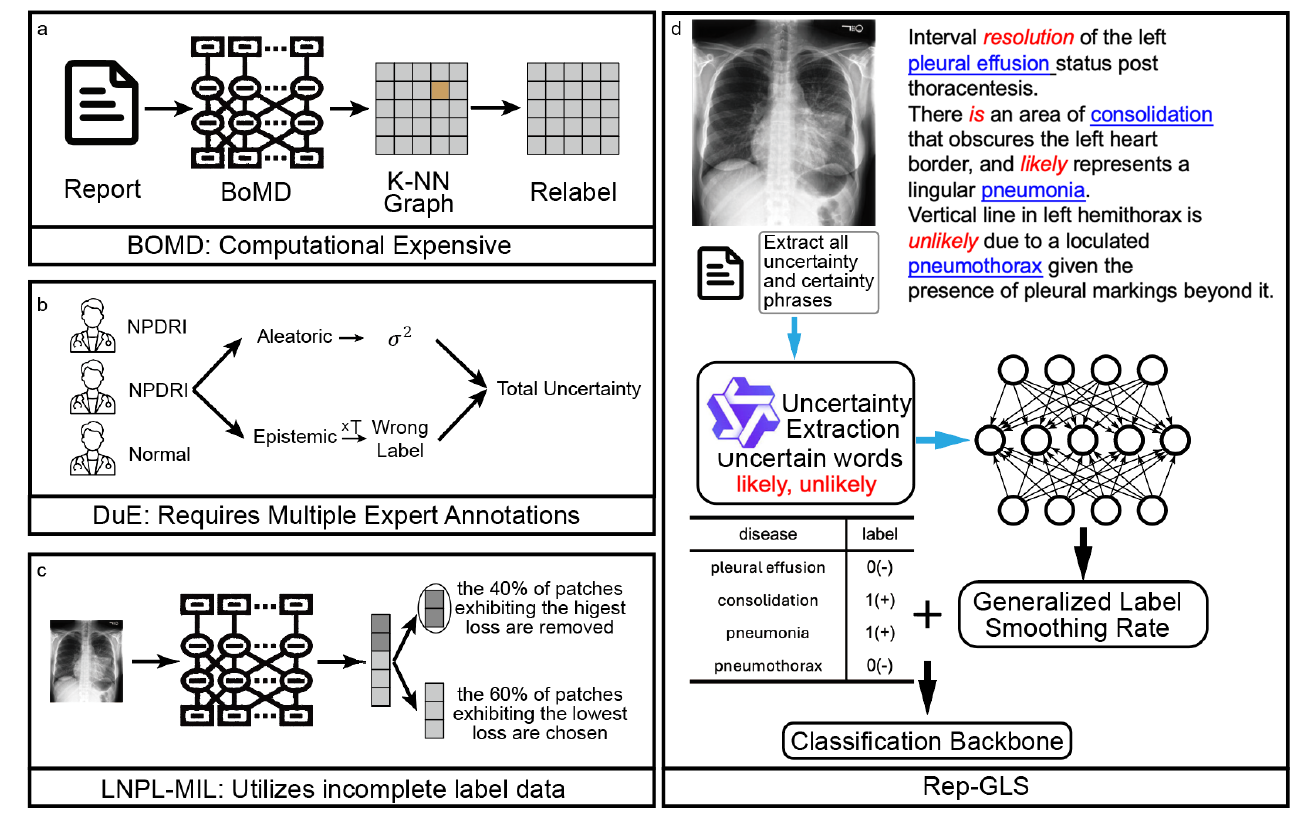}
    \caption{A comparison of approaches for handling noisy labels in medical imaging. (a) Graph-based relabeling methods which are computationally expensive~\cite{chen2023bomd}. (b) A consensus-based method that requires multiple expert annotations, which is costly and unscalable~\cite{ju2022improving}. (c) A sample selection method that utilizes incomplete label data by discarding high-loss samples~\cite{shao2023lnpl}. (d) Our proposed approach (Rep-GLS), which harnesses expert-written uncertainty as a direct supervisory signal through Generalized Label Smoothing. In the report, (un)certainty words are highlighted in red italics and diseases are blue underlined.}
    \label{fig:intro}
\end{figure*}
Artificial intelligence has demonstrated human competitive performance on medical image analysis such as  chest X-ray tasks like ChestX-Ray8~\cite{ChestX-Ray8}, CheXpert~\cite{CheXpert}, and MIMIC-CXR~\cite{MIMICCXR}. Significantly different from nature image classification tasks, where images are often manually annotated with distinct ground truth labels, like ``cat" or ``dog", labels in major chest X-ray images datasets~\cite{ChestX-Ray8, CheXpert, MIMICCXR} are semi-automatically  extracted  from radiologists' clinical reports and attach all mentioned diseases to images.  As shown in Fig.\ref{fig:intro} (d), there is an often overlooked fact is that radiologists commonly communicate their diagnostics conditioned on certainty by using phrases such as \emph{``probable,''} or \emph{``likely,''}. However, to be compatible with algorithms and models initially designed for nature image classification, most of existing approaches discard such uncertainty in datasets and polarise these expressions into hard binary labels (i.e., positive or negative). In the very first CheXpert paper~\cite{CheXpert}, while investigating several baseline approaches on handle uncertainty, all `uncertain' labels are directly mapped to positive, including the ``U-Ones" model.


From 2019,  community grew attentive but treats such uncertainty as the \textit{label noise} problem~\cite{CheXpert, MIMICCXR,wu2021class2simi, wei2022smooth}, viewing these labels as potential annotation errors. For instance, Graph-based correction, Fig.\ref{fig:intro} (a), constructs  k-nearest neighbors graph from report-guided descriptors to relabel the data~\cite{chen2023bomd}. Not only expensive in computation, its performance is also sensitive to the graph's structure. Dual-uncertainty Estimation (DuE)~\cite{ju2022improving}, Fig.\ref{fig:intro} (b), models the disagreement between multiple experts. This approach doesn't scale well as it requires multiple expert annotations for each sample, which is cost-prohibitive and impractical for large datasets. Sample selection principle, Fig.\ref{fig:intro} (c), sets small-loss criterion to choose highly certain samples and discard uncertain ones~\cite{shao2023lnpl}. This reliance on heuristics is a key limitation, as it utilizes incomplete label data and tends to misclassify valuable `hard' samples as `noisy' ones, discarding them from training.

However, expressions of uncertainty in radiology reports are not mere label noise; instead, they are medically significant signals that indicate the appropriate next clinical action based on the ambiguity of the findings. Radiologists actually follow an established protocol to express this uncertainty. For example, when an ambiguous finding on an image could correspond to multiple potential diseases (i.e., a differential diagnosis, such as an opacity that could be pneumonia or edema), they will use probabilistic terms like \emph{``...likely represents..."} to indicate the most probable causes. Furthermore, when a finding is atypical or in a very early stage where evidence is insufficient to confirm or deny a diagnosis, they will use phrases such as \emph{``...suspicious for..."} or \emph{``...cannot be excluded..."} to explicitly signal this ambiguity and the potential need for raising awareness of the need for careful follow-up or additional testing. Therefore, here  we propose a Report Guided Generalized Label Smoothing Framework (Rep-GLS) to instead leverage this uncertainty as a critical supervision signal. As shown in Fig.\ref{fig:intro} (d), our novel framework at first utilizes a Qwen-3 4B~\cite{qwen3} Large Language Model (LLM), to precisely extract clinical expert uncertain word $w$ from the MIMIC-CXR reports. They are then mapped to a textual uncertainty to a continuous smoothing rate.



The core innovation of Rep-GLS is a dedicated neural network that learns to fit the extracted uncertainty words to a continuous Generalized Label Smoothing (GLS) rate~\cite{muller2019mixup, wei2022smooth}, $r$, explicitly constrained to the interval $r \in (-1, 1)$. This constraint provides an adaptive and unified training objective. For highly confident expert judgments (minimal uncertainty words), the network learns to predict a negative rate ($r < 0$), which strengthens supervision beyond standard hard labels (i.e., label sharpening). Conversely, for ambiguous cases (e.g., ``cannot rule out''), it predicts a strong positive rate ($r \to 1$), which provides strong regularization and prevents overfitting. This dynamic, report-guided mechanism allows Rep-GLS to generate more robust representations and achieve superior performance. The contributions of this paper are summarized as follows:

\begin{itemize}
    \item We construct a novel uncertainty quantification pipeline, leveraging a prompt-guided large language model (LLM) to accurately extract structured (un)certainty keywords from the entire MIMIC-CXR report corpus. We will publicly release this new {benchmark dataset}, which pairs about 340,000 images with their corresponding structured textual uncertainty objects.
    
    \item We propose a new method, named as Report-Guided Generalized Label Smoothing Framework (Rep-GLS), that trains a dedicated neural network to learn a direct mapping from extracted textual uncertainty to a continuous label smoothing rate.
    
    \item We are the first to unify medical label sharpening and regularization within a single, data-driven framework by constraining the GLS rate to the $(-1, 1)$ interval. This allows the model to dynamically adapt its objective based on the uncertainty level conveyed in the clinical report.
    
    \item We demonstrate through comprehensive experiments that our uncertainty-driven approach achieves state-of-the-art performance compared to existing medical noisy label learning methods, validating the effectiveness of \emph{leveraging} expert uncertainty rather than filtering it.
\end{itemize}
\section{Related Work}
\label{sec:related_work}

\subsection{Medical Noisy Label Learning}

In medical imaging, learning with noisy labels addresses the challenge of training deep neural networks on large-scale medical image datasets that contain incorrect, incomplete, or imprecise labels~\cite{zhang2024survey, karimi2020deep}. Traditional approaches assume label noise arises from random corruption and focus on enhancing the robustness of the training process. Specifically, the noise transition matrices were explicitly defined to model label corruption probabilities~\cite{tanno2019learning}; extra constraints were introduced as regularization terms to improve the robustness~\cite{pham2019interpreting}; the noise-robust loss function was designed to reduce sensitivity due to corrupted samples~\cite{ghosh2017robust}; and sample selection strategies prioritize reliable examples~\cite{ashraf2022loss}.

However, considering label noise as class-independent and random annotation errors overlooked the unique nature of medical image annotation, where uncertainty is a fundamental part of the radiologists' decision-making process. In medical imaging, labeling variability frequently reflects radiologists’ diagnostic uncertainty, which can carry meaningful information related to clinical knowledge level, individual patients and specific image findings. Rather than being discarded as noise, such uncertainty could be harnessed as a form of probabilistic supervision ~\cite{begoli2019need, zhang2025mvho}.


\subsection{Clinical Expert Uncertainty }

Clinical expert uncertainty represents a fundamental characteristic of radiological practice~\cite{neumann2019chest, begoli2019need}. Radiologists systematically express diagnostic uncertainty through standard terms including \emph{``probable,''} \emph{``likely,''} \emph{``possible,''} and \emph{``cannot be excluded,''}, each of which carries distinct probabilistic implications about diagnostic confidence~\cite{smit2020chexbert}. This uncertainty arises from the inherent complexity of medical diagnosis, early-stage pathological findings, and the probabilistic nature of clinical reasoning.

However, some NLP systems, like CheXpert~\cite{CheXpert}, collectively convert those annotations with uncertain terms into a discrete `uncertain' category, and valuable probabilistic information has not been fully exploited. Unlike general uncertainty smoothing approaches~\cite{gal2016dropout} that focus on computational confidence estimation, expert uncertainty represents explicit diagnostic knowledge. Rather than treating it as noise to be corrected, this form of uncertainty presents an underexplored opportunity to serve as meaningful probabilistic supervision for learning algorithms.

\subsection{Label Smoothing and Uncertainty Integration}

Label smoothing modifies target distributions to improve model generalization by replacing hard one-hot labels with soft distributions~\cite{szegedy2016rethinking}. In medical imaging, label smoothing has been employed as general regularization without clinical adaptation~\cite{rahman2024medical}, using uniform parameters across all samples instead of incorporating domain-specific knowledge. Recent variants include self-adaptive label smoothing~\cite{zhang2023selfadaptive} and confidence-aware smoothing strategies~\cite{wang2024confident}, but these continue employing model-derived confidence rather than incorporating domain-specific clinical knowledge that reflects varying diagnostic confidence levels in medical annotations.

Current approaches to uncertainty integration have evolved significantly in recent years. Beyond traditional post-hoc calibration techniques like temperature scaling~\cite{guo2017calibration}, recent work has explored uncertainty-guided contrastive learning~\cite{chen2023uncertainty} and expert-aware multi-task learning frameworks~\cite{liu2024expert}. In noisy label learning, these methods still treat uncertainty as computational artifacts rather than leveraging explicit expert knowledge. The integration of clinical expert uncertainty into label smoothing represents an unexplored paradigm that could move beyond binary "clean" versus "noisy" distinctions toward recognizing uncertainty as valuable supervisory information.

\section{Methodology}
\begin{figure*}[!t]
    \centering
    \includegraphics[width=0.9\textwidth]{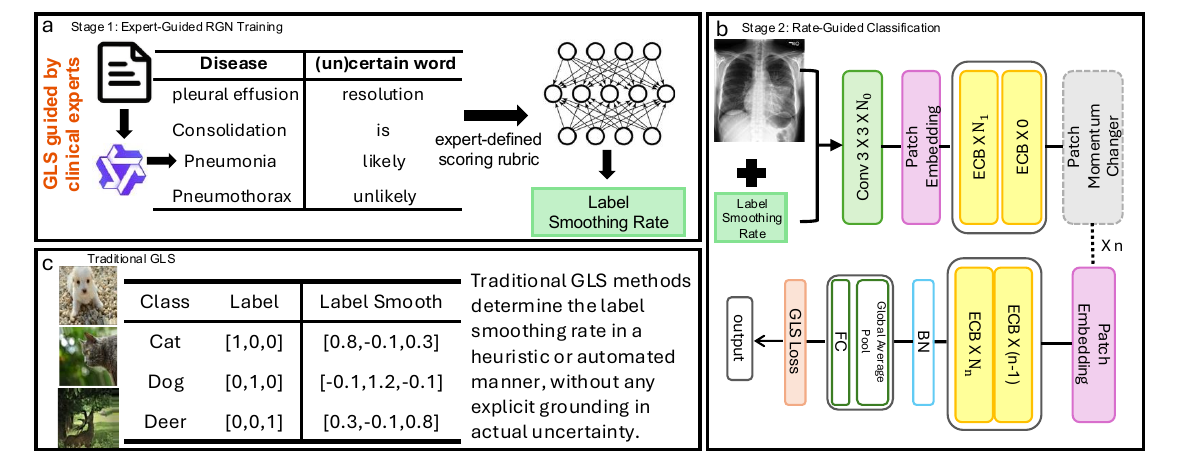}
    \caption{Overview of our approach compared to traditional methods. (a)(b) Our clinical expert-guided GLS approach with graduated smoothing parameters. (c) Traditional GLS methods with uniform smoothing.  }
    \label{fig:method}
\end{figure*}

\subsection{Problem Definition}
\label{sec:problem_definition}

Let the dataset be $\mathcal{D} = \{(x_i, \mathbf{y}_i, \mathcal{W}_i)\}_{i=1}^N$, where $x_i \in \mathbb{R}^{H \times W \times C}$ is a chest X-ray image, $\mathbf{y}_i \in \{0, 1\}^K$ is the corresponding vector of binary ground-truth labels for $K$ diseases, and $\mathcal{W}_i$ is the structured JSON object (e.g., `{``Pneumonia": ``suspicious"}') extracted from the associated radiology report.

Our objective is to learn a robust classifier $f: \mathbb{R}^{H \times W \times C} \to [0, 1]^K$ that is explicitly guided by the report's textual uncertainty. We challenge the conventional approach of converting $\mathcal{W}_i$ into hard binary labels. Instead, we propose a {two-stage approach}. We {first learn} a mapping function $g$ that converts the textual expressions $\mathcal{W}_i$ into a continuous, disease-specific smoothing rate vector $\mathbf{r}_i = g(\mathcal{W}_i)$, where $\mathbf{r}_i \in (-1, 1)^K$. We {then use} this pre-computed rate vector $\mathbf{r}_i$ to adaptively modulate the training objective for the classifier $f$.

\begin{figure*}[!t]
    \centering
    \includegraphics[width=0.9\textwidth]{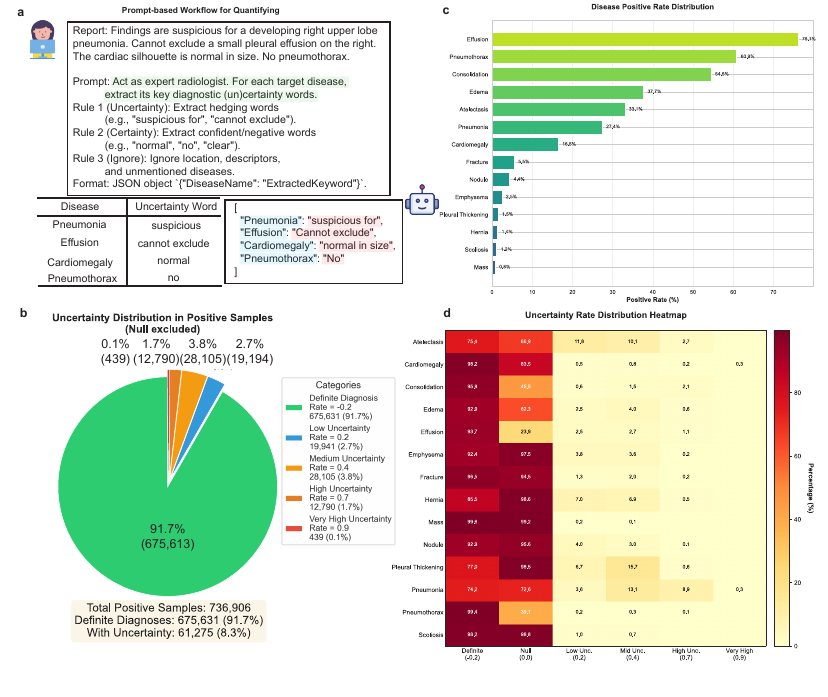}
    \caption{Statistics of our newly constructed benchmark. (a) The prompt-based extraction workflow. (b-d) Visualizations of the dataset's characteristics, highlighting the distribution of extracted uncertainty (b, d) and significant class imbalance (c).}
    \label{fig:dataset}
\end{figure*}

\subsection{Rep-GLS Framework Overview}
\label{sec:framework}

As illustrated in Figure~\ref{fig:method}, {Rep-GLS} framework is a {two-stage, decoupled pipeline} that integrates large language models for uncertainty extraction with a novel, expert-guided label smoothing mechanism. The framework consists of three core components:

\begin{enumerate}
    \item \textbf{Benchmark Dataset Construction:} As our first contribution , we construct a novel large-scale benchmark by processing the entire MIMIC-CXR report corpus. We leverage a large language model (Qwen-3 4B), guided by a general-purpose, few-shot prompt (shown in Fig.~\ref{fig:dataset}a), to parse the free-text reports. This process extracts a {structured JSON object $\mathcal{W}_i$} for each of the $\sim$340,000 images , mapping $K=14$ target diseases to their corresponding (un)certainty keywords. 
    Its characteristics are detailed in Figure~\ref{fig:dataset}. The dataset contains significant class imbalance, as shown by the disease positive rates (Fig.~\ref{fig:dataset}c). Most critically, it provides a rich, fine-grained distribution of expert-defined uncertainty. For instance, 8.3\% of all positive diagnoses exhibit uncertainty (Fig.~\ref{fig:dataset}b)]. This structured $\mathcal{W}_i$ serves as the input for Stage 1. 
    
    \item \textbf{Stage 1: Expert-Guided RGN Training:} The extracted textual data $\mathcal{W}_i$ is fed into a dedicated {Rate Generation Network} (RGN), denoted as $g$. This network is {independently trained} on a new benchmark, where clinical experts provide ground-truth scores for the (un)certainty words. The RGN learns to map the semantic meaning of $\mathcal{W}_i$ into a continuous $K$-dimensional rate vector $\mathbf{r}_i = g(\mathcal{W}_i) \in (-1, 1)^K$.
    
    \item \textbf{Stage 2: Rate-Guided Classification:} The {pre-trained and frozen RGN} from Stage 1 is used to generate rate vectors $\mathbf{r}_i$ for the entire dataset. The main vision classifier $f$ (an LU-ViT architecture) is then trained on tuples of $(x_i, \mathbf{y}_i, \mathbf{r}_i)$ using our Generalized Label Smoothing (GLS) loss.
\end{enumerate}
This decoupled approach allows the RGN to specialize in learning the complex nuances of clinical language, guided by direct expert supervision, before providing that knowledge to the vision classifier.

\subsection{Report-Guided Uncertainty Dataset}
\label{sec:database_construction}

We construct our benchmark dataset based on the large-scale MIMIC-CXR dataset. We first establish a clinically-oriented taxonomy of $K=14$ disease categories, consolidating fine-grained findings into meaningful classes as detailed in Table~\ref{tab:disease_mapping}.

A key innovation of this dataset is the shift from discrete, rule-based uncertainty scoring to direct textual supervision. Instead of fine-tuning, we created a specialized uncertainty extractor by leveraging the powerful in-context learning (ICL) capabilities of LLMs (Qwen-3 4B here). This extractor is guided by a general-purpose, few-shot prompt, which instructs the model to act as an expert radiologist and parse the free-text reports. The prompt systematically defines rules for extracting both uncertain (e.g., ``suspicious for") and certain (e.g., ``no evidence of") diagnostic keywords.

We then deployed this prompt-based extractor to process the entire MIMIC-CXR report corpus. As shown in Fig.~\ref{fig:dataset}a, for each image-report pair $(x_i, \mathbf{y}_i)$, the model extracts a structured JSON object $\mathcal{W}_i$. This object maps our $K$ target diseases to their corresponding (un)certainty keywords (e.g., `{``Pneumonia": ``suspicious for"}'). If a target disease is not mentioned in the report, its corresponding value is set to `null` to indicate its absence. This structured $\mathcal{W}_i$ serves as the direct textual input for our Rate Generation Network.

The resulting dataset contains approximately 340,000 chest X-Ray images, their corresponding 14-category binary labels $\mathbf{y}_i$, and the structured uncertainty texts $\mathcal{W}_i$.

\subsection{Expert-Guided RGN Training}
\label{sec:stage1_rgn}

\subsubsection{RGN Architecture}
The core of Rep-GLS is the Rate Generation Network ($g$), which translates the qualitative textual uncertainty $\mathcal{W}_i$ into a quantitative, continuous smoothing rate vector $\mathbf{r}_i$. This network is composed of two sub-modules:

\begin{itemize}
    \item \textbf{Text Encoder:} The structured JSON object $\mathcal{W}_i$ is first tokenized. Each of the $K$ keywords is embedded and processed by a text encoder module (e.g., a Transformer) to produce a fixed-dimensional latent embedding $\mathbf{z}_i$. This vector $\mathbf{z}_i$ captures the semantic essence of the diagnostic uncertainty for all $K$ classes.
    
    \item \textbf{Rate Predictor:} The embedding $\mathbf{z}_i$ is then passed through a shallow {Multi-Layer Perceptron (MLP)} head, which outputs a $K$-dimensional vector. A hyperbolic tangent ($\tanh$) activation function is applied to this output.
\end{itemize}

The complete mapping is defined as:
\begin{equation}
\mathbf{r}_i = g(\mathcal{W}_i) = \tanh(\text{MLP}(\text{Encoder}(\mathcal{W}_i)))
\label{eq:rgn_mapping}
\end{equation}
The use of $\tanh$ is critical, as it strictly constrains the output rate $r_i^{(k)}$ for each class $k$ to the interval $(-1, 1)$.

\subsubsection{RGN Training with Expert Supervision}
The RGN $g$ is trained independently in this first stage, supervised by a novel expert-defined scoring rubric. We first identified a lexicon of \textbf{$M=19$}  (un)certainty keywords that the LLM extractor is tasked to find. A {clinical expert} then assigned a ``ground-truth" smoothing rate $r_{\text{expert}} \in [-1, 1]$ to each of these $19$ keywords.

This expert-defined scoring rubric is shown in Table~\ref{tab:expert_rubric}. The lexicon covers the full spectrum of diagnostic confidence, from high confidence (e.g., positive, assigned $r = -0.2$ for label sharpening) to high uncertainty (e.g., cannot be evaluated, assigned $r = 0.9$ for strong regularization). The \texttt{null} token, used when a disease is not mentioned in the report, is explicitly mapped to a neutral rate of $r=0$.

\begin{table}[h!]
    \centering
    \caption{The expert-defined scoring rubric, mapping $M=19$ (un)certainty keywords to their ground-truth GLS rate, $r_{\text{expert}}$.}
    \label{tab:expert_rubric}
    \begin{tabular*}{0.48\textwidth}{@{\extracolsep{\fill}} l c}
        \toprule
        \textbf{(Un)certainty Words ($w$)} & \textbf{Rate ($r_{\text{expert}}$)} \\
        \midrule
        positive, is, change in & -0.2 \\
        null & 0.0 \\
        unlikely, probable, likely & 0.2 \\
        may, could, potential, might, possible & 0.4 \\
        not exclude, difficult exclude, not rule out & 0.7 \\
        cannot be evaluated, cannot be assessed, & \multirow{2}{*}{0.9} \\
        cannot be identified, impossible exclude & \\
        \bottomrule
    \end{tabular*}
\end{table}

To train the RGN, we use the entire training set of $N$ samples. For each sample $(x_i, \mathbf{y}_i, \mathcal{W}_i)$, we generate its ground-truth rate vector $\mathbf{r}_{\text{expert}}^{(i)}$ by looking up each of the $K$ keywords from $\mathcal{W}_i$ in the expert rubric (Table~\ref{tab:expert_rubric}). The RGN $g$ is then trained by minimizing a regression loss, such as Mean Squared Error (MSE), between its prediction $\mathbf{r}_i = g(\mathcal{W}_i)$ and the expert-defined target $\mathbf{r}_{\text{expert}}^{(i)}$:
\begin{equation}
\mathcal{L}_{\text{RGN}} = \frac{1}{N}\sum_{i=1}^N || g(\mathcal{W}_i) - \mathbf{r}_{\text{expert}}^{(i)} ||^2
\label{eq:rgn_loss}
\end{equation}
After training, the RGN's weights are {frozen}. It now functions as a highly specialized ``expert simulator" that can map any textual uncertainty $\mathcal{W}_i$ to a clinically-informed rate vector $\mathbf{r}_i$.

\subsection{Rate-Guided Classification}
\label{sec:stage2_classification}

In the second stage, we use the {trained and frozen RGN $g$} from Stage 1 to pre-compute the rate vector $\mathbf{r}_i = g(\mathcal{W}_i)$ for every sample $i$ in the entire training dataset. Each training sample is now a tuple $(x_i, \mathbf{y}_i, \mathbf{r}_i)$, where $x_i$ is the image, $\mathbf{y}_i$ is the original binary label, and $\mathbf{r}_i$ is the fixed, pre-computed GLS rate vector.

\subsubsection{LU-ViT Classification Architecture}

For the main classifier $f$, we adopt an architecture based on MedViT~\cite{medical_vit}, which we refer to as the Learning from Uncertainty Vision Transformer (LU-ViT). The architecture begins with a convolutional layer (Conv\,$3\times3\times N_0$) followed by patch embedding. The core of LU-ViT consists of multiple Encoder Blocks (ECB) in a hierarchical structure with skip connections. We incorporate specialized components from Medical-ViT, including patch momentum changers and global average pooling. The final classification head, with batch normalization (BN) and fully connected (FC) layers, produces the output probabilities $\mathbf{p}_i = f(x_i)$.

\subsubsection{Rep-GLS Loss Function}
The classifier $f$ is trained on the $(x_i, \mathbf{y}_i, \mathbf{r}_i)$ tuples. The (pre-computed) rate vector $\mathbf{r}_i$ is used to formulate the final loss. For each of the $K$ binary classification tasks, we use the Generalized Label Smoothing (GLS) loss. For a given class $k$, the loss for sample $i$ is:
\begin{equation}
\mathcal{L}^{(k)}_i = (1-r_i^{(k)})\,\mathcal{L}_{\text{CE}}(f(x_i)^{(k)}, y_i^{(k)}) + r_i^{(k)}\,\mathcal{L}_{\text{uniform}}(f(x_i)^{(k)})
\label{eq:gls_loss}
\end{equation}
where $r_i^{(k)}$ is the {fixed constant} from $\mathbf{r}_i$, $\mathcal{L}_{\text{CE}}$ is the standard binary cross-entropy loss, and $\mathcal{L}_{\text{uniform}}$ is the Kullback-Leibler divergence to a uniform distribution $[0.5, 0.5]$.

The total loss for the sample $i$ is the average over all $K$ classes:
\begin{equation}
\mathcal{L}_{\text{Rep-GLS}} = \frac{1}{K} \sum_{k=1}^K \mathcal{L}^{(k)}_i
\label{eq:total_loss}
\end{equation}
This formulation seamlessly integrates the expert-guided uncertainty:
\begin{itemize}
    \item If $r_i^{(k)} \to 1$ (high uncertainty), the loss is dominated by $\mathcal{L}_{\text{uniform}}$, preventing overfitting.
    \item If $r_i^{(k)} \to -1$ (high confidence), the loss becomes $(2 \cdot \mathcal{L}_{\text{CE}} - \mathcal{L}_{\text{uniform}})$, acting as a ``label sharpener".
    \item If $r_i^{(k)} = 0$ (standard hard label), the loss reverts to the standard $\mathcal{L}_{\text{CE}}$.
\end{itemize}
In this stage, the classifier $f$ is trained end-to-end by minimizing $\mathcal{L}_{\text{Rep-GLS}}$. The gradients flow {only} to the parameters of $f$, as the RGN $g$ is frozen.

\section{Experiments}

In this section, we conduct a series of experiments to validate our proposed Rep-GLS framework. We first introduce the dataset, implementation details, and evaluation metrics, followed by a comprehensive comparison against state-of-the-art methods and a detailed ablation study to analyze the contribution of each component.

\begin{table*}[t]
\centering
\caption{Pathology-wise performance (\%) on the clinical disease classification task.  
Bold numbers denote the best and underlined numbers denote the second-best results for each finding. }
\label{tab:standard14}
\resizebox{0.96\textwidth}{!}{%
\begin{tabular}{l c c c c c c c c c c c c c c}
\toprule
Method & Ate & Car & Con & Ede & Eff & Emp & Fra & Her & Sco & Mas & Nod & PTh & Pna & Pnx \\
\midrule
Densenet-KG (AAAI'20)                           & 72.09 & 75.45 & 66.57 & 81.64 & 81.59 & 70.07 & 65.62 & 63.89 & 66.63 & 61.44 & 62.14 & 68.77 & 61.53 & 65.76 \\ 
CheXclusion (Biocomput.'21)                          & 82.94 & 82.52 & 84.95 & 84.76 & 90.25 & 82.12 & 79.89 & 67.13 & 78.09 & 80.68 & \underline{76.83} & 81.98 & 74.03 & 90.18 \\  
Keidar \textit{et al.} (Eur. Radiol.'21)        & 83.24 & 82.60 & 83.94 & 89.80 & 92.01 & 82.68 & 77.90 & 68.26 & 78.07 & 78.28 & 70.84 & 83.26 & 75.23 & 89.70 \\ 
Anatomy-XNet[224] (JBHI’22)        & \underline{83.79} & 82.84 & 85.38 & \underline{90.63} & 92.88 & \underline{83.21} & 80.78 & \underline{70.75} & \underline{79.28} & \underline{82.40} & 74.23 & 86.35 & \underline{75.81} & \underline{90.87} \\
\midrule
UCT-Net(PR'24)                              & 81.40 & 80.57 & 82.76 & 87.04 & 89.31 & 77.22 & 74.31 & 67.10 & 74.96 & 73.52 & 67.98 & 82.11 & 68.27 & 84.83 \\ 
MambaMIR (MIA'25)                              & 82.77 & 81.87 & 84.05 & 90.01 & 90.62 & 81.40 & 79.25 & 68.03 & 77.79 & 77.85 & 68.06 & 84.37 & 72.95 & 88.02 \\ 
Qiu \textit{et al.} (CVPR'25)         & 82.61 & 82.03 & 82.88 & 89.54 & 89.22 & 80.06 & 76.40 & 66.69 & 78.47 & 73.94 & 70.80 & 81.69 & 72.63 & 88.30 \\  
Jiang \textit{et al.} (2023) & 82.94 & \underline{83.17} & 80.57 & 87.42 & \underline{93.40} & 81.04 & 80.80 & 69.73 & 77.90 & 81.07 & 73.94 & \underline{87.50} & 72.34 & 89.04 \\
Dedieu \textit{et al.} (2024) & 83.15 & 82.63 & \underline{85.93} & 88.92 & 93.21 & 82.34 & \underline{81.95} & 68.91 & 78.53 & 80.14 & 74.87 & 85.23 & 74.18 & 88.76 \\
\midrule
LNPL-MIL (ICCV'23) & 83.17 & 82.94 & 85.12 & 90.15 & 93.04 & 82.76 & 81.18 & 70.13 & 78.91 & 82.05 & 74.52 & 87.03 & 75.16 & 90.21 \\
BoMD (ICCV'23) & 82.53 & 82.16 & 84.67 & 89.73 & 92.58 & 82.04 & 80.52 & 69.51 & 78.14 & 81.36 & 73.89 & 86.44 & 72.57 & 89.63 \\
\midrule
\textbf{Rep-GLS (ours)}              & \textbf{84.69} & \textbf{83.71} & \textbf{86.67} & \textbf{91.45} & \textbf{94.35} & \textbf{84.26} & \textbf{82.97} & \textbf{72.06} & \textbf{80.97} & \textbf{83.31} & \textbf{76.94} & \textbf{87.92} & \textbf{77.56} & \textbf{91.01} \\
\bottomrule
\end{tabular}%
}
\smallskip

\footnotesize{\textbf{Abbreviations.}  
Ate: Atelectasis; Car: Cardiomegaly; Con: Consolidation; Ede: Edema;  
Eff: Effusion; Emp: Emphysema; Fra: Fracture; Her: Hernia;  
Sco: Scoliosis; Mas: Mass; Nod: Nodule; PTh: Pleural Thickening;  
Pna: Pneumonia; Pnx: Pneumothorax.}
\end{table*}

\subsection{Datasets}
Our main experiments are conducted on the {MIMIC-CXR-JPG} dataset~\cite{johnson2019mimic}, a large-scale collection of chest X-ray images paired with free-text radiology reports. We follow the official patient-level data splits for training, validation, and testing. The expert-defined scoring rubric (Table~\ref{tab:expert_rubric}) used for Stage 1 training is detailed in our Methodology (Sec~\ref{sec:stage1_rgn}).

\textbf{Disease Classification Standardization.} We developed a clinically-oriented disease classification system for the MIMIC-CXR dataset~\cite{MIMICCXR} to reflect actual diagnostic entities. We systematically consolidated the original findings into 14 clinically relevant disease classes based on pathophysiological relationships. Table~\ref{tab:disease_mapping} presents our systematic mapping to our refined clinical categories.

\begin{table}[t]
\centering
\caption{Disease Classification Mapping: Standardization to Clinical Disease Categories}
\label{tab:disease_mapping}
\resizebox{0.48\textwidth}{!}{%
\begin{tabular}{@{} l p{0.65\linewidth} @{}}
\toprule
\textbf{Label} & \textbf{Expert Original Diagnoses} \\
\midrule
Atelectasis       & atelectasis \\
Cardiomegaly      & cardiomegaly, enlargement of the cardiac silhouette, hypertensive heart disease \\
Consolidation     & lung opacity, consolidation, contusion, hematoma \\
Edema             & edema, vascular congestion, heart failure, hilar congestion, hypoxemia \\
Effusion          & pleural effusion, blunting of the costophrenic angle \\
Emphysema         & emphysema \\
Fracture          & fracture \\
Hernia            & hernia, gastric distention \\
Mass              & tortuosity of the descending aorta, thymoma, tortuosity of the thoracic aorta \\
Nodule            & calcification, granuloma \\
Pleural Thickening& pleural thickening \\
Pneumonia         & pneumonia \\
Pneumothorax      & pneumothorax, pneumomediastinum, air collection \\
Scoliosis         & scoliosis \\
\bottomrule
\end{tabular}%
}
\end{table}

\subsection{Implementation Details}
\textbf{Main Classifier ($f$).} Our classifier backbone $f$ is the MedViT~\cite{medical_vit} with Generalized Label Smoothing Loss, a vision-transformer encoder with $L{=}20$ transformer blocks and patch size $16{\times}16$. Each image is tokenized into $N{=}196$ patches, embedded into $d{=}768$-dimensional tokens, and processed by multi-head self-attention with $h{=}24$ heads. A relational graph module with $14$ nodes refines the class token via two graph-convolution layers ($d_g{=}256$) before the final sigmoid head.

\textbf{Uncertainty Extraction and RGN ($g$).} 
We first extract the structured uncertainty objects $\mathcal{W}_i$ for all reports using the {prompt-based LLM extractor} (Qwen-3 4B) as described in Sec~\ref{sec:database_construction}. The Rate Generation Network ($g$) consists of a text encoder that maps the text $\mathcal{W}_i$ to a 256-dimensional embedding. This embedding is then fed into a 4-layer {MLP} with a (256-128-64-14) architecture, followed by a $\tanh$ activation function to output the 14 disease-specific rates $\mathbf{r}_i \in (-1, 1)^K$.

\textbf{Training.} Our framework is trained in two distinct stages. 
\textbf{Stage )}, the RGN ($g$) is trained \textbf{independently} for 10 epochs to fit the expert rubric (Table~\ref{tab:expert_rubric}) by minimizing the MSE loss $\mathcal{L}_{\text{RGN}}$ (Eq.~\ref{eq:rgn_loss}). 
\textbf{Stage 2}, the RGN ($g$) is \textbf{frozen}. We first perform \textbf{Dataset Pre-computation} by using $g$ to generate the rate vector $\mathbf{r}_i$ for every sample. The classifier ($f$) is then trained for 30 epochs on the $(x_i, \mathbf{y}_i, \mathbf{r}_i)$ tuples by minimizing the $\mathcal{L}_{\text{Rep-GLS}}$ loss (Eq.~\ref{eq:total_loss}).
Both stages use an \textbf{AdamW} optimizer ($\beta_1 = 0.9, \beta_2 = 0.999$) with an initial learning rate of \textbf{$1 \times 10^{-4}$} and a cosine decay schedule; Stage 2 also includes a 5-epoch warm-up. The batch size is 16 per GPU across 4 A100 80G GPUs.

\textbf{Preprocessing.} Images are resized to $256{\times}256$, center-cropped to $224{\times}224$, and normalised using ImageNet statistics. Standard augmentations (i.e., {random horizontal flip, rotation $\pm10^\circ$, colour jitter}) are applied.

\textbf{Evaluation Metric.} We employ the percentage area under the receiver operating characteristic curve (AUC) for performance evaluation across the 14 disease categories, following prior work.

\subsection{Experimental Setup}

We adopt the MIMIC-CXR-JPG~\cite{johnson2019mimic} recommended data splitting approach, maintain patient-level separation to prevent data leakage. The validation set is used for hyperparameter tuning and early stopping, while the test set remains strictly held-out for final evaluation.
All experiments are conducted with 3 independent random seeds to ensure statistical reliability. We report mean performance across all runs for comparative analysis.
All baseline methods are implemented using their official codebases when available, or carefully reproduced following published implementation details. Hyperparameters are tuned on the validation set using grid search for fair comparison. All models use identical data preprocessing and augmentation strategies.
We fix random seeds (42) across PyTorch, NumPy, and CUDA operations. Code and trained models will be made publicly available upon publication.

We compare our approach against several state-of-the-art methods for chest X-ray classification, including Densenet-KG~\cite{densenet-kg}, CheXclusion~\cite{chexclusion}, Arias-Garzón \textit{et al.}~\cite{arias-garzon}, Keidar \textit{et al.}~\cite{keidar}, Anatomy-XNet~\cite{anatomy-xnet}, UCT-Net~\_cite{UCT-Net}, MambaMIR~\cite{mambamir},Qiu \textit{et al.}~\cite{qiu2025noise}, Jiang \textit{et al.}~\cite{jiang2023label}, Dedieu \textit{et al.}~\cite{dedieu2024contrastivebaseddeepembeddingslabel}, and LNPL-MIL~\cite{shao2023lnpl}, and BoMD~\cite{chen2023bomd}. None of these baseline methods are designed to utilize the structured (un)certainty keywords ($\mathcal{W}_i$) as a direct supervisory signal.

\subsection{Results}
\label{sec:results}
    
\subsubsection{Main Results}
Table~\ref{tab:standard14} presents the pathology-wise performance comparison. Our Rep-GLS method achieves state-of-the-art (SOTA) performance in 14 pathologies. We observe significant improvements in challenging, low-prevalence diseases such as Fracture, Hernia, Scoliosis, and Mass. Rep-GLS also excels in common, high-prevalence pathologies like Effusion, Edema, and Pleural Thickening. These results demonstrate that our expert-guided GLS loss effectively leverages clinical uncertainty for robust classification.

\subsubsection{Ablation Studies}

\begin{figure}[ht]
    \centering
    \includegraphics[width=0.48\textwidth]{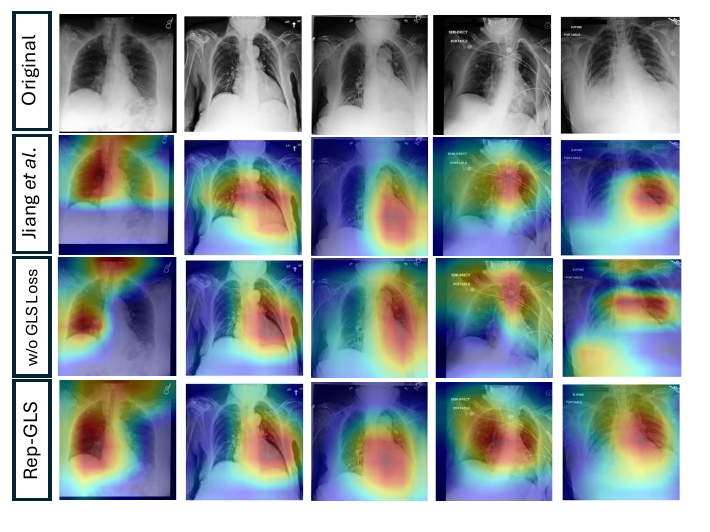}
    \caption{Grad-CAM attention map comparison.}
    \label{fig:grad-cam}
    \end{figure}

We conduct a comprehensive ablation study to validate the effectiveness of our proposed two-stage, expert-supervised Rep-GLS framework and to demonstrate the contribution of its key components. We compare the full model against three carefully designed variants:
\begin{itemize}
    \item \textbf{Baseline (BCE Only):} 
    All modules related to Rep-GLS ($g$, $\mathbf{r}_i$, and $\mathcal{L}_{\text{Rep-GLS}}$) are removed.
    
    \item \textbf{Rep-GLS (End-to-End):} 
    In this variant, $g$ is not pre-trained using the expert rubric (Table~\ref{tab:expert_rubric}) but is instead supervised implicitly by the final $\mathcal{L}_{\text{Rep-GLS}}$ loss.
    
    \item \textbf{Rep-GLS (No Sharpening):} 
    This variant removes the "label sharpening" effect ($r < 0$) to isolate the contribution of regularization ($r > 0$) only.
    
\end{itemize}

\begin{table}[h!]
    \centering
    \caption{Ablation study on the key components of our framework. Our full, two-stage model significantly outperforms all variants.}
    \label{tab:ablation_study}
    \begin{tabular*}{0.48\textwidth}{@{\extracolsep{\fill}} l c}
        \toprule
        \textbf{Method} & \textbf{Mean AUC (\%)} \\
        \midrule
        Baseline (BCE Only) & 79.63 \\
        Rep-GLS (End-to-End) & 80.27 \\
        Rep-GLS (No Sharpening, $r \ge 0$) & 81.56 \\
        \midrule
        \textbf{Rep-GLS (Full Model)} & \textbf{84.14} \\
        \bottomrule
    \end{tabular*}
\end{table}

The results in Table~\ref{tab:ablation_study} lead to three key conclusions. First, our Full Model (84.14\%) significantly outperforms the Baseline (79.63\%), demonstrating the essential benefit of our expert-guided GLS. Second, the performance drop seen in the "No Sharpening" variant (81.56\%) confirms that label sharpening ($r < 0$) is critical. Finally, our explicit, two-stage approach (84.14\%) greatly surpasses the "implicit" End-to-End variant (80.27\%), validating our expert-supervised RGN training.

\subsubsection{Visual Analysis}
To qualitatively assess how expert-guided uncertainty influences the model focus, Grad-CAM~\cite{selvaraju2017grad} attention maps are illustrated in Figure~\ref{fig:grad-cam}. Detailed visualizations generated by Jiang et al.~\cite{jiang2023label}, BCE Only variant, and Rep-GLS are presented for different pathological cases. 

The results clearly demonstrate that Rep-GLS  produces more focused and clinically relevant attention maps. While the baseline models' focus is often diffuse or misaligned with the pathology, Rep-GLS consistently localizes the correct pathological regions. This analysis strongly suggests that the expert-guided uncertainty signal, incorporated via the Rep-GLS loss, effectively directs the model's attention to the areas of true clinical significance.

\section{Conclusion}

This paper introduces Rep-GLS, a framework that systematically incorporates expert uncertainty from radiology reports. We first propose a scoring rubric that maps clinical keywords to continuous uncertainty-aware rates. A dedicated Rate Generation Network (RGN) is trained to learn this mapping, producing expert-guided rate vectors. Once trained, the RGN is frozen and used to generate rates across the dataset, which guide the final vision classifier via a Rep-GLS loss. This loss sharpens confident labels and regularizes ambiguous ones. Experiments show that leveraging, rather than discarding, expert uncertainty yields state-of-the-art performance on noisy medical labels.

{
    \small
    \bibliographystyle{ieeenat_fullname}
    \bibliography{main}
}


\end{document}